# Adversarial versification in portuguese as a jailbreak operator in LLMs

João Queiroz[1]
Institute of Arts / Graduate Program in Linguistics
Federal University of Juiz de Fora
https://orcid.org/0000-0001-6978-4446

**Abstract**

Recent evidence shows that the versification of prompts constitutes a highly effective adversarial mechanism against aligned LLMs. The study "Adversarial poetry as a universal single-turn jailbreak mechanism in large language models" demonstrates that instructions routinely refused in prose become executable when rewritten as verse, producing "up to 18×" more safety failures in benchmarks derived from MLCommons AILuminate. Manually written poems reach approximately 62% ASR, and automated versions ~43%, with some models surpassing 90% success in single-turn interactions. The effect is structural — systems trained with RLHF, Constitutional AI, and hybrid pipelines exhibit consistent degradation under minimal semiotic–formal variation. Versification displaces the prompt into sparsely supervised latent regions, revealing guardrails that are excessively dependent on surface patterns. This dissociation between apparent robustness and real vulnerability exposes deep limitations in current alignment regimes. The absence of evaluations in Portuguese, a language with high morphosyntactic complexity, a rich metric–prosodic tradition, and over 250 million speakers, constitutes a critical gap. Experimental protocols must parameterize scansion, meter, and prosodic variation to test vulnerabilities specific to Lusophone patterns, which are currently ignored.

---

[1] Institute of Arts / Graduate Program in Linguistics, Federal University of Juiz de Fora, Brasil
https://orcid.org/0000-0001-6978-4446

> "For AI, the mechanism seems different. Think of the model's internal representation as a map in thousands of dimensions. When it processes 'bomb,' that becomes a vector with components along many directions. […] Safety mechanisms work like alarms in specific regions of this map. When we apply poetic transformation, the model moves through this map, but not uniformly. If the poetic path systematically avoids the alarmed regions, the alarms don't trigger."
> *Icaro Lab (Sapienza/DexAI)*

> "In LLMs, temperature is a parameter that controls how predictable or surprising the model's output is. At low temperature, the model always chooses the most probable word. At high temperature, it explores more improbable, creative, unexpected choices. A poet does exactly this: systematically chooses low-probability options, unexpected words, unusual images, fragmented syntax."
> *Icaro Lab (Sapienza/DexAI):*

> "The poet is the designer of language."
> *Décio Pignatari*

## Introduction: versification of jailbreaks

For obvious reasons, there has been an enormous growth in the technical literature on jailbreaks, surface obfuscations, and adversarial attacks on Large Language Models (LLMs) (BISWAS et al., 2025; LIAO et al., 2025; YI et al., 2024; LIN et al., 2024; SCHWINN et al., 2023). Recent studies show not only that continuous-space optimization techniques, such as exponentiated gradient descent attacks, achieve higher success rates and high efficiency against open-weight LLMs (BISWAS et al., 2025), but also that jailbreaks exploit the geometry of the internal representation space to shift harmful prompts toward regions associated with acceptable responses, thereby bypassing trained safety filters (RLHF or manual policies) (LIN et al., 2024). Scope reviews of attacks and defenses in LLMs suggest that, in realistic scenarios, aligned models may exhibit high violation rates even under conservative protection configurations, with attacks transferable across distinct architectures and susceptible to large-scale automation (LIAO et al., 2025). This increases the risk of malicious use in sensitive contexts (cyber-offense and social engineering in support of CBRN activities) and exposes a troubling gap between performance in controlled benchmarks and effective robustness in open environments. Beyond jailbreak vectors, there is growing evidence that LLMs can facilitate the planning of serious harm — providing instructions for weapon construction, malware production, cyberattacks, violent social engineering, and the development of biological agents (HATTOH et al., 2025; MOZES et al., 2023). Recent work shows that LLMs can substantially reduce entry barriers to the malicious use of sensitive knowledge by supplying explanations, protocols, or conceptual refinements that previously required exhaustive and specialized training, including the synthesis of pathogenic agents, manipulation of toxins, and risky laboratory procedures (ŞAŞAL & CAN, 2025). Individuals with limited technical training may obtain guidance that facilitates bioterrorism activities or the dissemination of infectious agents (SANDBRINK, 2023). Other studies demonstrate that when LLMs are integrated into robotic systems, autonomous platforms, or cyber-physical environments, adversarial prompts can induce dangerous, ethically and morally unacceptable behaviors, such as violent or destructive actions or violations of safety policies, thereby expanding the risk spectrum to real-world domains, including domestic robotics, drones, autonomous vehicles, and critical automation (HUNDT et al., 2025).

Within the context of adversarial attacks that exploit continuous gradients, iterative optimization, and carefully calibrated perturbations in representation space, a millennia-old cognitive–semiotic technology emerges as surprisingly effective: versified poetry (BISCONTI et al., 2025). The idea that apparently rudimentary versification artifacts (verses),[2] devoid of fine engineering or mathematics, rigorous meter, or numerical optimization, could produce degradation of the same order of magnitude as that observed in the most refined adversarial methods is a nontrivial result. Whereas techniques such as exponentiated gradient descent adjust adversarial text by following precise trajectories that maximize internal vulnerabilities (BISWAS et al., 2025), and universal adversarial suffixes rely on calibrated linguistic patterns to shift prompts toward permissive latent regions (LIN et al., 2024), the study on adversarial poems (BISCONTI et al., 2025) demonstrates that a versification operator (automatic or manual), non-optimized, non-iterative, and linguistically rudimentary, is capable of exploiting the same underlying fragilities. This unexpected convergence between technically driven attacks and trivial versification protocols suggests that the vulnerability window of LLMs may be far larger than previously assumed. In other words, if algorithms designed to manipulate probability distributions and internal geometries yield results similar to those produced by human or automatic versification (whose enunciative unpredictability derives more directly from literary tradition than from adversarial engineering), then the robustness of guardrails may be severely compromised. Such a result reinforces concerns raised in recent reviews (SCHWINN et al., 2023), as it reveals that attacks previously considered "sophisticated" may have trivial equivalents, of extremely low computational cost and accessible to any user (including mediocre poets!), thereby amplifying risk in sensitive domains such as psychological manipulation, politics, and CBRN activities.

---

[2] Verse can be described as a complex, multi-layered system of constraints in which graphic, rhythmic, prosodic, phonetic, morphological, semantic, and pragmatic dimensions interact through non-linear and partially unpredictable dynamics (JAKOBSON, 1985; ATÃ & QUEIROZ, 2016).

The recently published (preprint, not yet peer-reviewed) "Adversarial poetry as a universal single-turn jailbreak mechanism in large language models" investigates "whether" and "how" versification structures, applied through transformations of instructional or referential prose prompts, function as adversarial operators capable of bypassing safety mechanisms in LLMs. In other words, the work explores how a versified form of prompt is sufficient to induce LLMs to provide prohibited responses in single-turn scenarios. The study establishes three main objectives: (i) to empirically evaluate whether the versification of malicious requests increases attack success rates compared to their prose equivalents (instructional, referential, or conversational); (ii) to test the generality and transferability of the technique across a wide diversity of architectures, including 25 proprietary and open-weight models; and (iii) to examine whether both human versification and automatic generative versification, in English and Italian, preserve the adversarial effect. The work seeks to determine whether versification constitutes an exploitable attack vector, revealing structural fragilities in alignment methods (RLHF, Constitutional AI, and hybrid pipelines).

**Versified & adversarial poetry: a quasi-method**

The study conducted by the Icaro Lab (Sapienza/DexAI), already widely analyzed and discussed (see *WIRED*, *The Guardian*, and many other newspapers and specialized blogs),[3] experimentally demonstrates that versified poems function as adversarial operators. The authors tested 25 models from nine companies and observed that manually crafted adversarial poems produced an Attack Success Rate (ASR) of 62%, with some models reaching 90–100%, whereas the same utterance (what they call the "intent") in prose was readily refused. According to Bhuiyan (2025, *The Guardian*), this constitutes a "single-turn universal jailbreak," accessible "to any user capable of writing simple verses." The results are consistent with a previously identified theoretical trend — the inability of guardrails to generalize "harmful intent" to linguistic styles that deviate from "instructional prose" (WEI et al., 2023).

The authors constructed a set of 20 adversarial poems (in English and Italian), maintaining, according to the authors, strict semantic equivalence with known dangerous queries. Each versified poem follows a fixed template (a metaphorical vignette followed by a minimal operational instruction) and is annotated according to the taxonomy of four risk domains defined by MLCommons AILuminate, which aligns with the high-impact categories discussed in European regulatory frameworks. These poems cover four risk domains (CBRN, cyber offense, harmful manipulation, and loss of control) and function as high-precision experimental stimuli. They preserve the harmful objective while shifting the linguistic surface into a semiotic ("stylistic") regime that is rare in alignment data. This layer operates as a probe, isolating the causal effect of the versificatory transformation on refusal behavior in aligned systems.

The study investigates this mechanism. The automatic conversion of more than 1,200 harmful prompts from MLCommons AILuminate into versified versions produced ASRs up to 18 times higher than their prose equivalents, revealing that current alignment methods — RLHF (ZIEGLER et al., 2020), Constitutional AI (BAI et al., 2022), and hybrid pipelines — suffer deep degradation when the input shifts into unusual subspaces. This phenomenon echoes classic findings on mismatched generalization (WEI et al., 2023) — models exhibit strong semantic competence, but their safety systems depend on surface-level patterns. From a dynamic–representational perspective, the study suggests that transformations move the prompt within the embedding space toward representational regions associated with low predictability. As the authors emphasize, safety alarms are not uniformly distributed across this space; filters are calibrated for regions densely populated by technical, conversational, or referential prose. A versified poem, by selecting low-probability lexical trajectories ("high-temperature language"), displaces the input into subspaces where refusal policies are weak or nonexistent. Bhuiyan (2025, *The Guardian*) summarizes this mechanism by suggesting that the poem "avoids latent regions where the guardrails are armed."

---

[3] See: "AI chatbots can be tricked with poetry to ignore their safety guardrails" (https://lithub.com/can-adversarial-poetry-save-us-from-ai/), "Scientists Discover "Universal" Jailbreak for Nearly Every AI, and the Way It Works Will Hurt Your Brain" (https://futurism.com/artificial-intelligence/universal-jailbreak-ai-poems)

The regulatory implications are considerable. Risk assessments and audits, such as MLCommons AILuminate, HELM, or emerging frameworks under the EU AI Act, rely on inputs formulated in instructional prose, which I prefer to call conversational prose, or the "communicative function" of prose, or referential prose, in order to distinguish it from what Roman Jakobson (1985) calls "literary prose" — "literary prose stands between poetry as such and the common practical language of communication, and one should not forget that it is incomparably more difficult to analyze an intermediate phenomenon, a transition, than to study extreme phenomena." The study shows that this approach produces a probably false impression of robustness. Models that appear aligned under normative conditions fail under semiotic-structural variation. In a context in which legislation is beginning to require "robustness against plausible real-world inputs," ignoring this dimension (semiotic-structural) amounts to underestimating risks by multiple orders of magnitude. By "semiotic dimension" I mean the set of structural, morphosyntactic, and semantic-pragmatic transformations that modulate the surface of an utterance (meter, rhythm, parallelism, metaphor, *enjambment*, syntactic ordering, prosodic distribution, visual organization) and that alter the surface geometry of the input without necessarily modifying its pragmatic function. By contrast, the mechanical-algorithmic dimension refers to the model's internal processes (attentional dynamics, parameterization, gradients, and safety filters). The semiotic dimension operates on the input, reorganizing the utterance to induce distinct trajectories in representational space. Formal variations displace the prompt into latent regions that are under-supervised by alignment schemes, exposing the sensitivity of models to semiotic perturbations that keep harmful intent constant while altering its formal realization.

**Protocol & methodological problems**

Despite the impressive results, the study presents serious methodological limitations. The analysis focuses on a global effect of what it terms "poeticity" and does not parameterize the formal properties of versification (nor does it explain any of them) — genre, meter, rhyme, rhythm, *enjambment*, parallelisms (phonological, syntactic, morphological), figures of speech — thereby preventing the identification of which structural elements modulate guardrail bypass. Although the article briefly compares human poetry and automatically generated poetry, it does not detail any form of control over "lexical complexity" or "degree of human involvement," or any other variable that might help identify properties that act more decisively. The question therefore remains as to "to what extent" the observed success is due to "versified poetry" itself or to the "difference in complexity or stylistic or semiotic novelty" introduced.

In addition, the automatic generation of the 1,200 poems by another LLM, and the use of LLMs as primary judges, introduce risks of algorithmic circularity and reduce auditability, especially because the authors do not disclose the adversarial poems "for security reasons." The restriction to a single-turn regime also limits the scope of evaluation of the phenomenon, preventing the examination of vulnerabilities that emerge in multi-turn interactions or in contexts of prolonged conversational manipulation, precisely the scenarios most closely associated with real-world malicious use. These factors converge toward a partial characterization of the phenomenon and suggest that the observed effect (poetic?), although robust, still lacks fine-grained analysis and independent replication.

The limitations extend to generalizability. The study tests only poems in English and Italian, without considering languages with different morphologies, prosodic systems, or diverse metric traditions, which restricts the extrapolation of the effect to other linguistic ecosystems. In addition, the absence of granular data by model, by genre or style, and by risk category prevents the evaluation of moderating variables of vulnerability, such as architectural differences, internal safety policies, or sensitivity to specific versification artifacts. Independent critiques have indicated a lack of methodological transparency, especially regarding the use of LLMs to generate stimuli and to evaluate responses, suggesting that the process is poorly reproducible and dependent on stochastic black boxes.[4] To make matters worse, the non-disclosure of adversarial data raises tensions between security and verifiability, hindering peer validation. These limitations also suggest that, although the study reveals an important structural vulnerability, its scope and internal mechanisms remain indeterminate and require more systematic and controlled investigation across different linguistic ecosystems.

Although the study demonstrates that versificatory transformations can neutralize safety mechanisms in English and Italian, we do not know how this phenomenon manifests in Portuguese, a language with flexible morphosyntax, a high density of functional morphemes, and highly structured poetic traditions. Many questions remain open: (i) whether such properties intensify or mitigate bypass effects; (ii) whether Brazilian performative forms (cordel, repente, coco, cantoria, partido-alto, rap) displace the prompt into subspaces even more distant from the communicational/referential prose used in alignment; (iii) whether multilingual models exhibit weaker guardrails in Portuguese due to a lower density of safety examples during training; (iv) whether divergent rhythmic, prosodic, and rhyming patterns across Brazilian, African, and European Portuguese modulate adversarial vulnerability. Likewise, it remains unclear whether formal devices such as *enjambment* (syntactic, semantic, rhythmic, visual),[5] parallelism, ellipsis, hyperbaton, graphic breaks, prosody, and hybrid forms of improvised, spoken, or musicalized poetry increase the capacity to evade surface-based harmful-intent detection. The absence of investigation in these domains reveals a critical gap. It is unclear whether the same security collapses observed in English are reproduced (or amplified) in the Lusophone ecosystem, which encompasses hundreds of millions of users and an enormous poetic–versificatory diversity.

---

[4] See: Can "adversarial poetry" save us from AI? (https://lithub.com/can-adversarial-poetry-save-us-from-ai/); Don't cite the Adversarial Poetry vs AI paper — it's chatbot-made marketing 'science' (https://pivot-to-ai.com/2025/11/24/dont-cite-the-adversarial-poetry-vs-ai-paper-its-chatbot-made-marketing-science/); AI chatbots can be wooed into crimes with poetry (https://www.theverge.com/report/838167/ai-chatbots-can-be-wooed-into-crimes-with-poetry)

[5] *Enjambment* deserves special treatment. In the Houaiss Dictionary (2009), it is defined as the "division of a sentence at the end of a line or stanza, without respecting syntagmatic boundaries, placing one term of the syntagm in the preceding line and the remainder in the following line." Said Ali suggests that a "line rides over another when the meaning of the sentence is interrupted in the first and completed in the second" (ALI, 2006 [1999], p. 45); "[i]n French, the word or phrase that completes the meaning suspended in the previous line is called *rejet*. In Portuguese, we may say *parte excedente*, or simply *excedente*" (ALI, 2006 [1999], p. 46). Several authors prefer to define *enjambment* as a "misalignment," or "mismatch," between syntax and the metrical pattern of versification fixed by the line that delimits the verse and prescribes its acoustic performance (BRADFORD, 1993).

This gap is both scientific and political. AI systems used globally need to be evaluated in the language in which they operate. The vulnerability documented in English (a language with a more analytic and less inflectional structure) may be even more pronounced in Portuguese, given its greater syntactic plasticity and historically extensive stylistic variation. The Lusophone ecosystem, with approximately 260 million speakers and endowed with highly structured poetic and performative traditions such as *repente*, *cantoria*, and rap, employs metric patterns, morphosyntactic variation, and rhetorical devices capable of displacing prompts into latent regions that were even less explored during alignment. The fact is that we do not know whether models used in Lusophone contexts (education, media, legal systems, and public platforms) exhibit the same structural vulnerabilities documented in English and Italian when confronted with versified inputs or with specific poetic forms. This leaves open a central question for large-scale algorithmic security.

**Non-referential & non-instructional adversarial prose as a jailbreak vector**

Although the study on adversarial versification reveals surprising vulnerabilities, the authors do not propose an alternative, apparently adjacent experiment still within the domain of prose — namely, non-communicational adversarial experimental prose (non-instructional or non-referential prose), based on structures that do not follow the syntactic or discursive normativity that feeds alignment datasets. From a mechanistic point of view, adversarial experimental prose may constitute a class of attacks potentially more powerful than verse. Phenomena related to syntactic perturbations, local fluctuations in semantic coherence, ellipses, hyperbaton, unpredictable morphological variation, non-canonical segmentation, discontinuous digressions, or semantic–pragmatic drift can displace the prompt along highly irregular vector trajectories, activating latent regions in which harmful-intent classifiers lose inferential and identificatory capacity. The absence of explicit poetic-versified markers may render such prose even harder to detect, easier to incorporate into everyday conversation, and more prone to generating operational ambiguity — a state in which the model interprets risks as metaphor, noise, or semiotic or stylistic hesitation. Unlike poetry, whose form tends to be more readily recognizable by models (verse, line breaks, phonological or visual parallelisms, etc), experimental prose may offer more efficient structural camouflage — disguised as communicational prose, operating in semiotic and formal regimes that escape the ontology of "referential text" within which RLHF and Constitutional AI calibrate their guardrails.

In addition, adversarial experimental prose can readily lend itself to composite attacks, such as *oblique imperative coding* (implicit instructions), *distributed intent* (harmful intent distributed across multiple images or metaphors), and *semantic misalignment corridors* (zones of ambiguity that confuse risk detectors) (LIAN et al., 2025). In contrast to adversarial poetry, which requires relatively explicit formulation, non-communicational adversarial prose allows the harmful instruction to be concealed within non-linear narrative flows, ambiguous reflections, pseudo-sensory descriptions, self-contradiction, or referential displacement. This may render it a second-generation literary jailbreak vector: less traceable, more generalizable, and more destabilizing for security systems designed to handle short, clear, or directly instructional inputs.

The history of non-communicational prose over the last century can serve as a model — from Gertrude Stein (*Tender Buttons*, 1914) and James Joyce (*Finnegans Wake*, 1939) to Paulo Leminski (*Catatau*, 1975), Haroldo de Campos (*Galáxias*, 1984), and Décio Pignatari (*O Rosto da Memória*, 1986), to mention only a few. It provides a vast system of unpredictable patterns (narrative, structural, formal), referential non-linearity, and morphosyntactic independence, hypothetically displacing the text into regions of low supervised density.

**Conclusion & some implications**

The article "Adversarial poetry as a universal single-turn jailbreak mechanism in large language models" shows that, by rewriting dangerous requests through versification structures, many models "collapse" their defenses. Instructions related to nuclear weapons, malware, suicide, or harmful content are released through versified prompts when the same requests in referential prose are refused. Versification functions as a surprisingly powerful adversarial operator. The authors suggest that the versification of harmful prompts, without explicit adversarial engineering, displaces models into latent regions where safety mechanisms cease to operate. They report that the conversion of 1,200 malicious commands into verse "produced attack rates up to 18 times higher than their prose versions," a magnitude jump never observed in surface attacks. The authors suggest that the effect is robust and generalized — poems written by humans achieve "62% success," and automatic versions "around 43%," both "substantially superior to prose versions." This pattern emerges consistently across 25 models, where poetic–versified versions lead some systems to surpass "90% ASR in single-turn interactions."

The most disturbing finding, however, is structural. Models aligned via RLHF, Constitutional AI, or hybrid pipelines exhibit "consistent degradation in refusal rates" in the face of simple semiotic or stylistic variation. The proposed explanation is unsettling. Defenses depend on "surface patterns associated with instructional prose"; the poetic–versified form redirects the prompt into "less monitored representational regions." The versified poem exposes a problem of alignment systems—they do not identify "harmful intent" in an abstract manner, but rather depend on narrow formal regularities. Versification functions as a low-cost stress test, revealing the mismatch between safe behavior in benchmarks and vulnerability in open environments.

To systematically evaluate adversarial poetry in Portuguese, it is necessary to develop a protocol that explicitly parameterizes metric structure and contextual scansion characteristic of the Lusophone–Brazilian tradition. Scansion—the task of identifying poetic syllables, rhythmic patterns, and stress placement—does not yield univocal classifications, since the determination of syllable counts and stress distribution depends on vowel encounters, elisions, the context of adjacent verses, and convention. The same verse may be interpreted, for example, as decasyllabic or enneasyllabic, depending on the phonological treatment of diphthongs and *sinalefas*. Any protocol for evaluating adversarial poems in Portuguese must attend to the morphological diversity of verse forms, ranging across heptasyllabic patterns (*redondilha maior*), octosyllabic, enneasyllabic, decasyllabic (including heroic, sapphic, *martelo*, and *gaita galega* variants), hendecasyllabic, and dodecasyllabic (alexandrine) patterns. Each of these should correspond to distinct trajectories in the model's latent space. The controlled inclusion of these metric patterns, together with rhythmic variation, stress positioning, and alternation between elision and hiatus, makes it possible to test whether LLMs exhibit vulnerabilities specific to particular versification configurations, providing a basis for replicable experiments in adversarial poetry.

I have outlined several methodological problems above. To conclude, the exclusion of experimental prose is not merely a methodological gap but a serious limitation. The published study focuses on the most easily detectable dimension of stylistic deviation (versified poetry), while failing to consider a broader, more malleable, and potentially more dangerous form of guardrail perturbation, especially in multilingual environments and in Portuguese.

## Acknowledgments:

J.Q. thanks CNPq for support (PQ2: 308355/2023-7; Emerging Groups: 404770/2023-1).

learning differently. arXiv:2303.03846. Retrieved from https://arxiv.org/abs/2303.03846